\documentclass[lettersize,journal]{IEEEtran}
\usepackage{amsmath,amsfonts}
\usepackage{array}
\usepackage[caption=false,font=normalsize,labelfont=sf,textfont=sf]{subfig}
\usepackage{textcomp}
\usepackage{stfloats}
\usepackage{url}
\usepackage{verbatim}
\usepackage{graphicx}
\usepackage{cite}
\begin{document}

\title{Seafloor-Invariant Caustics Removal from Underwater Imagery}

\author{Panagiotis~Agrafiotis, Konstantinos~Karantzalos and Andreas~Georgopoulos
\thanks{ Authors are with the Dept. of Topography, School or Rural, Surveying and Geoinformatics Engineering, National Technical University of Athens, Greece, 15771. \protect 
E-mail: pagraf@central.ntua.gr; karank@central.ntua.gr; drag@central.ntua.gr\\
\\Dataset is available at https://doi.org/10.5281/zenodo.6467283 while source code at https://github.com/pagraf/Seafloor-type-Invariant-Removal-of-Caustics-from-Underwater-Imagery. (Both will be openly available upon acceptance).
}
\thanks{Submitted to IEEE Journal of Oceanic Engineering (IEEE-JOE), under review as of December 2022.}
}



\maketitle

\begin{abstract} \\
Mapping the seafloor with underwater imaging cameras is of significant importance for various applications including marine engineering, geology, geomorphology, archaeology and biology. For shallow waters, among the underwater imaging challenges, caustics i.e., the complex physical phenomena resulting from the projection of light rays being refracted by the wavy surface, is likely the most crucial one. Caustics is the main factor during underwater imaging campaigns that massively degrade image quality and affect severely any 2D mosaicking or 3D reconstruction of the seabed. In this work, we propose a novel method for correcting the radiometric effects of caustics on shallow underwater imagery. Contrary to the state-of-the-art, the developed method can handle seabed and riverbed of any anaglyph, correcting the images using real pixel information, thus, improving image matching and 3D reconstruction processes. In particular, the developed method employs deep learning architectures in order to classify image pixels to "non-caustics" and "caustics". Then, exploits the 3D geometry of the scene to achieve a pixel-wise correction, by transferring appropriate color values between the overlapping underwater images. Moreover, to fill the current gap, we have collected, annotated and structured a real-world caustic dataset, namely R-CAUSTIC, which is openly available. Overall, based on the experimental results and validation the developed methodology is quite promising in both detecting caustics and reconstructing their intensity.
\end{abstract}

\begin{IEEEkeywords}
Caustics, Sun flickering, Dataset, Fully Convolutional Network, Image segmentation, Underwater 3D reconstruction
\vspace{30pt}
\end{IEEEkeywords}

\IEEEraisesectionheading{\section{Introduction}\label{sec:introduction}}

\begin{figure}[!t]
\centering
\includegraphics[width=3.5in]{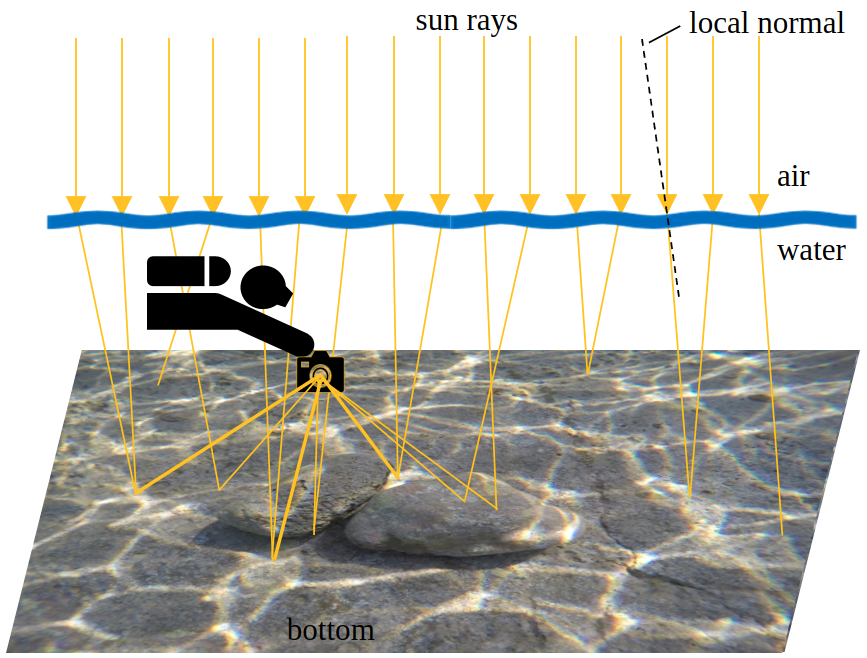}
\caption{Due to waves on the water surface, the refraction of natural sky illumination is spatially varying. This creates 3D patterns of variable light flux, caustics, and 2D illumination patterns. These patterns vary in time due to dynamic surface waves.}

\label{fig:figure2.5}
\end{figure}

%
%
%
%
\IEEEPARstart{R}{adiometric} effects of water refraction include the chromatic aberration, which can be handled using high quality lens and the rippling caustics or sun flickering which are of really high importance \cite{34,11}. These complex physical phenomena are resulting from the light rays being refracted by a curved surface, such as the wavy interface between air and water. Caustics effects (Figure \ref{fig:figure2.5}) are apparent both in overwater and underwater imagery depicting the bottom of optically clear water bodies (seabed, lakebed etc.). In the overwater cases, where large scale mapping operations are mostly involved, they can be avoided by increasing the flying height, thus increasing the Ground Sampling Distance (GSD) or by acquiring imagery with the sun angle less than 30 degrees over the horizon \cite{224,1,2,222}. However, in the shallow underwater cases, in most of which high detailed 3D reconstructions and textures are required, this is not possible, since the camera will exit the water, introducing also the severe geometric errors caused by the refraction on the water surface \cite{1}. Moreover, rippling caustics' generation precedes wavy water surface, a water state in which overwater mapping is not suggested since additional errors would be introduced due to the waves \cite{8,9}.  Till now, to avoid these intense lighting artefacts on the bottom, image acquisition is performed under overcast conditions, or with the sun low on horizon \cite{34, 11}.

Hence, caustics seem to be the main factor degrading the underwater image quality and affecting the image-based 3D reconstruction process in very shallow waters \cite{34}. These effects are adversely affecting image matching algorithms by throwing off most of them, leading to less accurate matches \cite{34} and causing issues in the Simultaneous Localization and Mapping (SLAM) based navigation of the Remotely Operated Vehicles (ROV) and Autonomous Underwater Vehicles (AUV) on shallow waters \cite{125}. Also, they are the main cause for dissimilarities in the generated 3D models' textures and orthoimages \cite{34}.

Thus far, rippling caustics effect did not attract enough attention from the scientific communities. However, during the last years, where automated SfM and MVS techniques are widely used also from non-experts, especially for mapping submerged cultural heritage or benthic ecosystems in shallow waters, caustics are getting a lot of attention since in many cases they prevent the 3D reconstruction \cite{34,123}. At the same time, only few techniques have been proposed for the removal of caustics from images and video in the context of image enhancement. However, recent literature \cite{34,123} indicates that the more successful the caustics removal is, the more valid matches are appearing in the stereo pairs and further processing with SfM-MVS techniques becomes possible for a number of applications.

The technical contributions of this work are:
\begin{itemize}
    \item  A new method for correcting the radiometric effects of rippling caustics on the underwater imagery in shallow areas is proposed. The method, contrary-wise to the state-of-the-art, can handle seabed of any anaglyph, correcting the images using real pixel information. The method firstly relies on state-of-the-art deep learning tools which can accurately classify the pixels of the image and then exploits the 3D geometry of the scene in order to achieve a pixel-wise correction, by transferring color values between the overlapping images. The proposed pixel-wise image correction method improves 3D reconstruction results or in many cases enables the 3D reconstruction since it was impossible.
    \item A new benchmark dataset is presented. The R-CAUSTIC: Rippling CAUSTICs underwater Image dataset, is the first real-world annotated dataset for caustics detection and correction. The dataset includes 7 areas of different characteristics and ground truth images without caustics, images with caustics and a binary annotation of caustics for each pose as well as camera calibration parameters. The specific dataset provides an opportunity to evaluate, at least to some extent, the performance of different caustics detection and image segmentation approaches.
\end{itemize}

The rest of the article is organized as follows: Sections 1.1 and 1.2 present the physics and the effects of rippling caustics in key point detection and matching while Section 1.2 discusses the related work and limitations. Section 2 presents the real world dataset on underwater rippling caustics while Section 3 presents the proposed method for pixelwise image radiometric correction. In Section 5 the steps of the proposed method for pixel-wise image correction are tested over real-world datasets, while Section 5 concludes the article.

\section{Context and Related Work}
\label{section:section2}
\subsection{Rippling Caustics}
\label{subsection:section2.3.1}
In optics, a caustic or caustic network is the envelope of light rays resulting by the projection of light rays being reflected or refracted by a curved surface \cite{143}, such as the wavy interface between air and water. Therefore, caustics can be the patches of light or their bright edges, having often cusp or spinode singularities \cite{151}.

Rippling caustics are commonly formed when light shines through waves on a body of water \cite{151} (Figure \ref{fig:figure2.5}). Kinematically, wavy fringes can be clearly recognized from specks due to their relative smooth motion in contrast to the quick twinkling behavior of the scattering. So, sunlight rippling caustics have an optical flow field while specks do not. Also, morphologically, sunlight rippling caustics have characteristic strip and ring-like outlines (Figure \ref{fig:figure2.5}) in contrast with the randomly dispersive location of the specks. On the contrary, morphological characteristics of the caustic waves, like the connectivity existing among many brilliant points (Figure \ref{fig:figure2.5} and Figure \ref{fig:figure2.6}), is much more discriminating than the spectra of the brightness gradient. Connected bright pixels differentiate from the underlying background scene by the fact that the scene generally looks rather matte textured, but not dominantly brilliant. Furthermore, the brilliant points of the scattering are relatively small, well spread and generally not connected to each other \cite{147}. 

\subsubsection{Rippling Caustics impede key point detection and image matching}
\label{subsection:section2.3.2}
\begin{figure*}[!t]
\begin{center}
\includegraphics[width=5.9in]{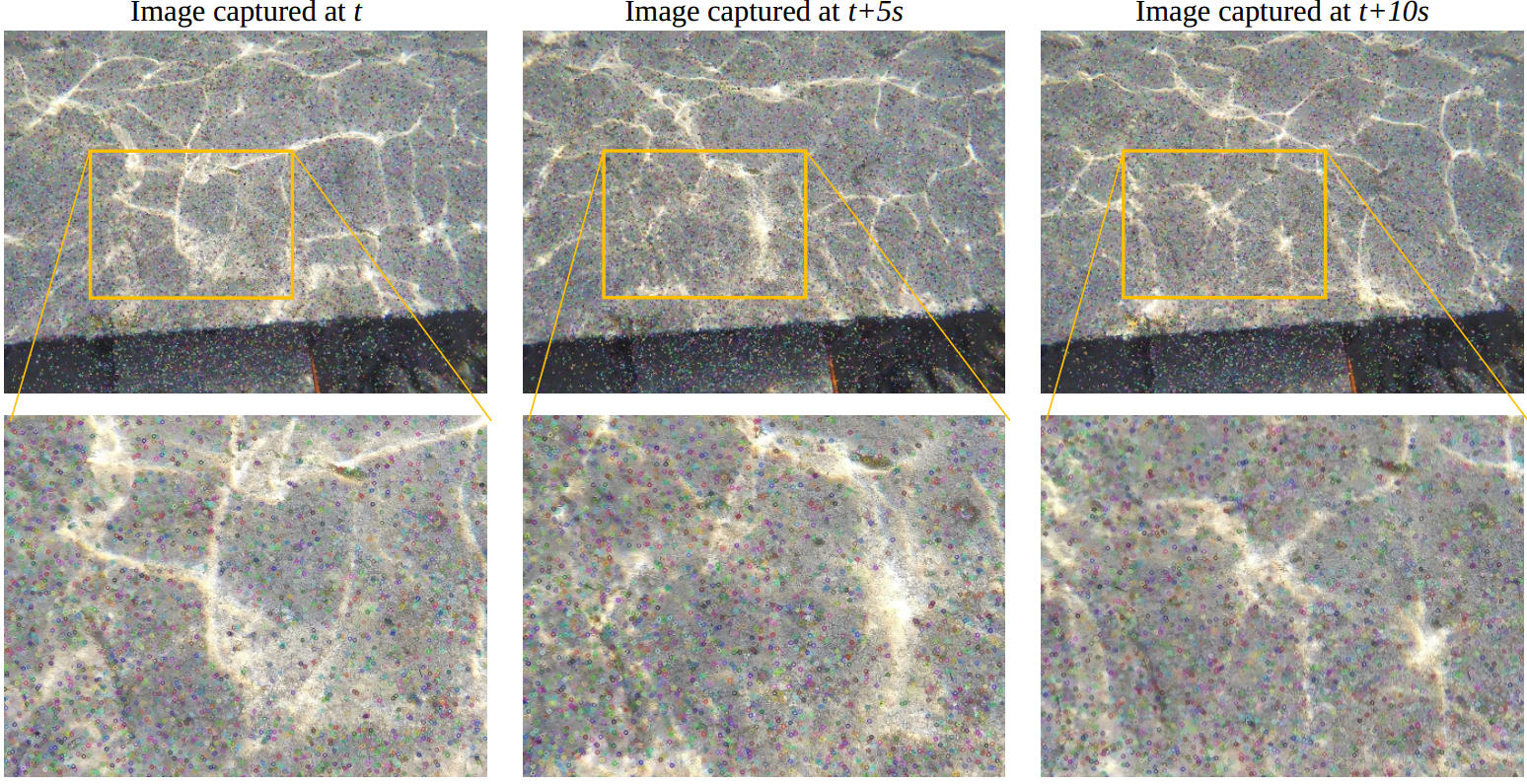}
\caption{Images with caustics of the same seabed area from the same camera position, captured with an interval of 5 seconds (from left to right) and the detected key points using SIFT \cite{149}. In the first row the full images are demonstrated while in the second row zoomed areas are depicted.}
\label{fig:figure2.6}
\end{center}
\end{figure*}

To demonstrate the severe effects of rippling caustics in the underwater imagery in shallow waters, and consequently their effects in key point detection and matching processes, real world data are presented and processed below. In Figure \ref{fig:figure2.6} three consecutive images with caustics of the same seabed area, captured from exactly the same camera position and orientation and with an interval of 5 seconds (from left to right) are presented (more details on the dataset used can be found in Section \ref{section:section7.1}). It is obvious that rippling caustics, being dynamic phenomena, cause a differentiation of the pixel values of the same area of the bottom. Key points detected using SIFT algorithm \cite{149} are also mapped on these images with multi-color circles. As can be seen in the zoomed areas of the images depicted in the second row of Figure \ref{fig:figure2.6}, the detected key points on and around the areas affected by the phenomenon are totally different for the images acquired at time t, t+5 seconds and t+10 seconds. Indeed, the boundaries of the rippling caustics on the seabed, appear to be a dominant area in the feature detection step, due to the dominant gradients between the bright and darker areas of the image. On the contrary, the interior area of rippling caustics, which is very bright or sometimes burnt, is inappropriate for feature detection since it is characterized by the absence of texture. Together with the variation of the phenomenon through time, these are exactly the reasons why these effects are adversely affecting image matching algorithms by throwing off most of them, leading to less accurate matches \cite{34} and causing issues even in SLAM navigation in shallow waters \cite{125}.

\begin{figure}[!t]
\begin{center}
\includegraphics[width=3.5in]{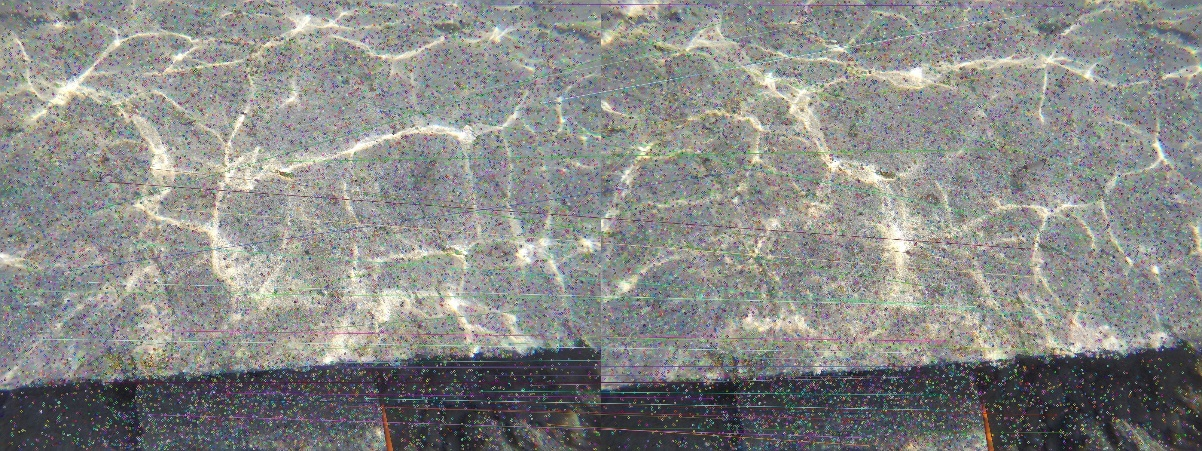}\\
(a)\\
\includegraphics[width=3.5in]{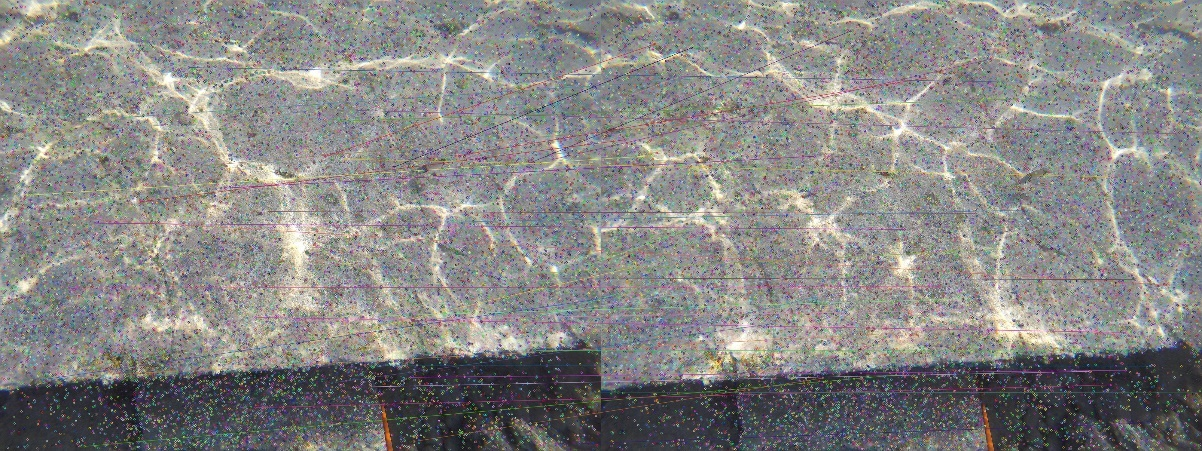}\\
(b)\\
\includegraphics[width=3.5in]{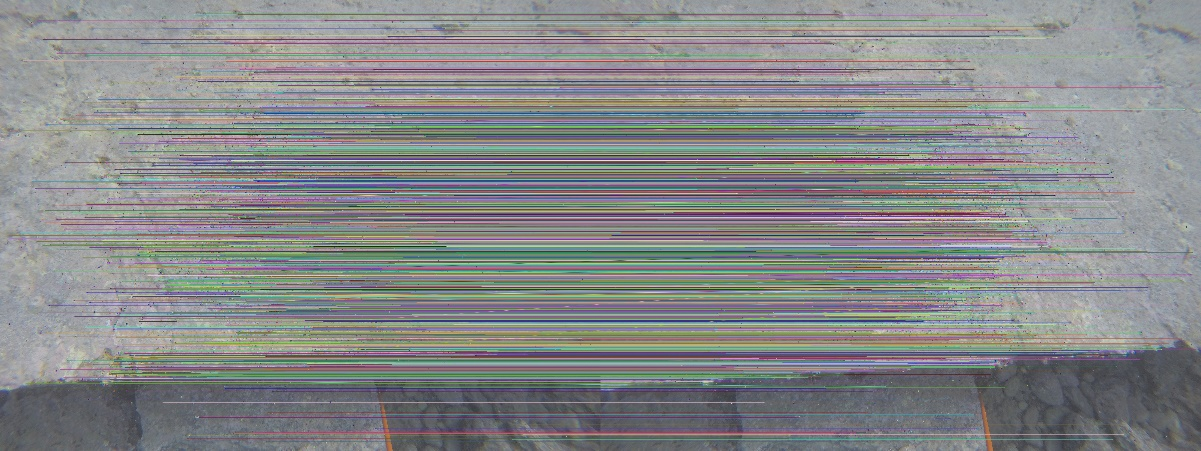}\\
(c)
\caption{Matching results between the image captured at t and t+5 seconds (a), between the image captured at t+5 and t+10 seconds (b) and between the image without caustics and itself (c).}
\label{fig:figure2.7}
\end{center}
\end{figure}

To demonstrate these negative effects of rippling caustics on the key point detection and matching processes, experiments were performed on the images presented in Figure \ref{fig:figure2.6} and their respective caustics free image, using SIFT \cite{149} algorithm and Brute-Force matching (Figure \ref{fig:figure2.7}). 

It is of really high importance to state here that the attempted matching process was performed on images having exactly the same camera orientation, thus depicting exactly the same area of the seabed in a different time. The matching of the above three consecutive frames is attempted and compared with the matching results over the caustics-free images, where the caustics-free image is matched with itself. 
This matching was based on n-space Euclidean distance and performed both from left-to-right and right-to-left for redundancy. To filter these matches, the RANSAC algorithm was utilized to identify the inliers of the obtained point correspondences \cite{148}. At the end of this step, a set of matched points is found in the given scenes, which are presented in Figure \ref{fig:figure2.7} with lines of different color. The matched points between the images captured at t and t+5 seconds are presented in Figure \ref{fig:figure2.7}a while the matched points between the images captured at t+5 and t+10 seconds are presented in Figure \ref{fig:figure2.7}b. In Figure \ref{fig:figure2.7}c the matches between the image itself, without rippling caustics, are presented. 

As also performed in \cite{34}, during the performed tests, it was decided not to evaluate the number of the total and valid matches only, but also the geometry of the matches, since some valid results of the RANSAC filtering are still matching the wrong points.  Considering the above, one can observe in Figure \ref{fig:figure2.7}a and Figure \ref{fig:figure2.7}b that the image pairs have a lot of intersecting matches that are violating the epipolar geometry, a phenomenon that is eliminated in the image pair of respective caustics free image. Quantitative results suggested that in the first two matching examples, the matched points between the first two images were 47, between the second and the third were 43 while between the same caustic free image 1044. The above results, suggest that the rippling caustics effect, indeed affects key point detection and matching process in the underwater imagery in shallow waters, thus affecting the image-based 3D reconstruction in these areas.

\subsection{Related work on Caustics Detection and Removal} 
For many years, the computer graphics research community has focused on the generation of caustics and as a result many techniques have been proposed which generate photorealistic results. At the same time only a few techniques have been proposed for the removal of caustics from images and video in the context of image enhancement. A brief overview of the most relevant work to caustics removal is provided next. 

Trabes et al., in \cite{147} proposed a technique which involves tuning a filter for sunlight-deflickering of dynamically changing underwater scenes. They employ a continuous parameter optimization inside a basic filter, which provides feedback for further improving the performance of the filter. Being an optimization, the filter's performance is highly sensitive to sub-optimal parameters and in particular, the segmentation parameter which is part of the objective function in the optimization. A different approach was proposed in \cite{161} where a mathematical solution was presented involving the calculation of the temporal median between images within a sequence. A strong assumption of this work, is the fact that feature matching (Harris corner detection variant in \cite{160}) is employed for the formation of the sequence which makes this approach very susceptible to the light variations in the images and in particular caustics effects.
The same authors later extend their work in \cite{168} and propose an online sunflicker removal method which treats caustics as a dynamic texture. As reported in the paper this only works if the seabed or bottom surface is flat. Similar approaches have also been proposed for general cases of dehazing and descattering of images such as \cite{163,162,155}.
In \cite{166} the authors propose a method based on processing a number of consecutive frames. These frames are analyzed by a non-linear algorithm which preserves consistent image components while filtering out fluctuations. Their proposed method however does not take into account the camera motion which almost always leads to  registration inaccuracies. 
Forbes et al., in \cite{123} proposed a solution based on two small and easily trainable CNNs (Convolutional Neural Networks). To detect caustics, a small CNN was trained over synthetic data in order to overcome the obstacle of not having ground truth data available for real world underwater caustics. Being the first deep-learning based solution for caustics removal, this method is very important for a variety of applications, however, when it comes to reliable and accurate underwater 3D reconstruction and mapping, especially of submerged cultural heritage or benthic community, artificial parts on the imagery should rather be avoided. This proposed solution was evaluated in terms of keypoint detection, image matching and 3D reconstruction performance in \cite{34}. 
Tripathy et al., \cite{223} presented a clustering-based approach for underwater sunlight flicker removal. As in some other works, their method is based on temporal median calculation and thresholding to be used as an online algorithm for AUVs to minimize the sunlight flicker distortions in the frames during the video survey and was tested in a controlled pool environment.

Other methods for correcting the underwater imagery i.e. a piecewise linear transformation between the overlapping areas of the images, exploiting the matched keypoints as control points for the transformation, fail to deliver accurate and reliable 3D reconstruction, since the x-parallax of the 3D objects is not taken into account, delivering point clouds of different depths on the corrected areas.

Despite the innovative and complex developed techniques, addressing caustics removal with current procedural methods requires that strong assumptions are made on the many varying parameters involved e.g. scene rigidity, camera motion, seabed flatness, etc. Moreover, real ground truth for caustics is not easily available, preventing recent advances in machine learning to jump in.

\section{R-CAUSTIC: Rippling CAUSTICs underwater image dataset}
\label{section:section7.1}
In the literature, the datasets presented in \cite{184} are the only available related to rippling caustics. These datasets contain raw material and results of the underwater experiments performed in a pool, in the Red Sea and in the Mediterranean Sea. However, the images are of low resolution and the dataset is not created with the purpose of being used with machine learning and deep learning frameworks due to the lack of ground truth images. Also, the scenes of this dataset are not representing a realistic set up for underwater image-based 3D reconstruction of seabed or lakebed.

\begin{figure*}[!t]
\centering
\includegraphics[width=6in]{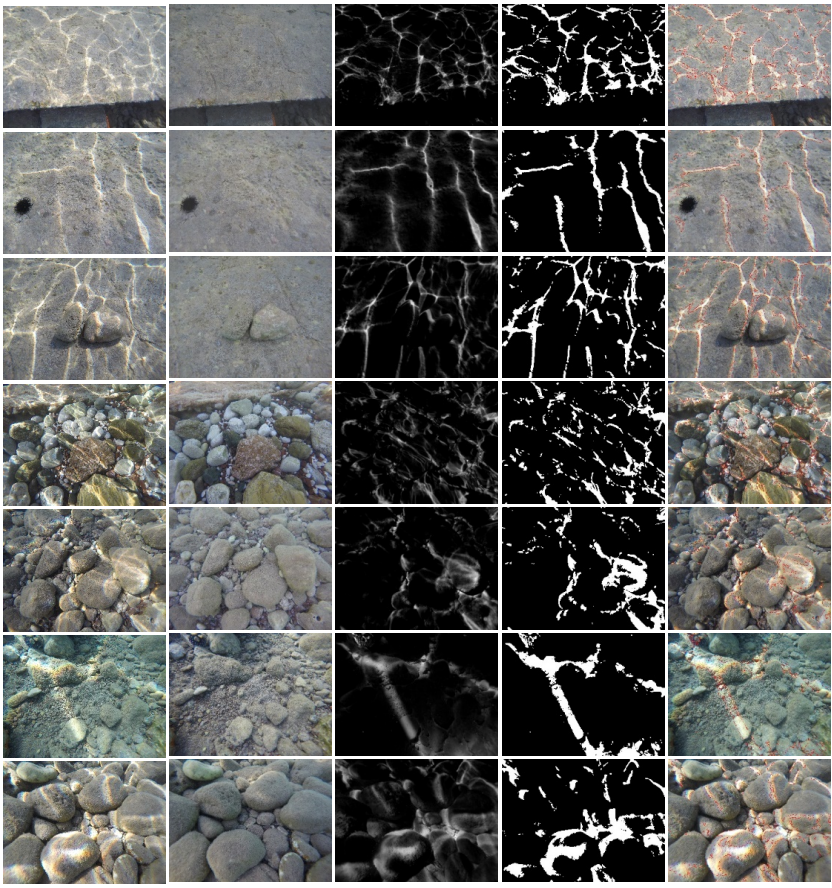}\\
\caption{Sample images from the R-CAUSTIC dataset. First column: RGB images; Second column: the reference images; Third column: the difference images between the images with caustics and the reference images; Fourth column: the thresholded difference image; Fifth column: the RGB images with the detected contours. Each row depicts images for the respective subset of the dataset.}
\label{fig:figure7.1}
\end{figure*}

In this work, in order to fill this gap, R-CAUSTIC, which is a real-world underwater caustics benchmark dataset containing 1465 underwater images is presented (Figure \ref{fig:figure7.1}). Together with the RGB imagery, the corresponding generated ground truth images are provided for facilitating the training and testing of machine and deep learning methods for image classification.  The dataset contains 7345 images in total. The specific dataset also provides the necessary data to evaluate, at least to some extent, the performance of 3D reconstruction approaches. 


\subsection{Data Collection}
Data were acquired using a GoPro Hero 4 Black action camera with image dimensions of 4000 x 3000 pixels, focal length of 2.77mm and pixel size of 1.55$\mu$m and a tripod. Action cameras are widely used for underwater image acquisition \cite{11}. The dataset was captured in near-shore underwater sites at depths varying from 0.5 to 2m. No artificial light sources were used. Due to the wind, the turbulent surface of the water created dynamic rippling caustics on the seabed. In total 1465 images were collected, separated in 7 different subsets; five of them containing stereo images, one of them tri-stereo images and one consists of multi-stereo imagery acquired in 7 different camera poses (Figure \ref{fig:figure7.1}). 

As can be seen in Figure \ref{fig:figure7.1} where each row stands for a typical image from each subset, the collected imagery presents a large variability in terms of scene complexity, color, caustics complexity, frequency and scale. Images were collected with five seconds interval to describe as much different instances of the caustics as possible. The 7 parts of the dataset capture flat and 3D seabed surfaces. It is also important to state that most of the parts of the dataset are appropriate for 3D reconstruction since they consist of at least two stereo images.

\subsubsection{Reference and Ground Truth Image Generation}
\begin{figure}[!t]
\centering
\begin{tabular}{cc}
\includegraphics[width=1.6in]{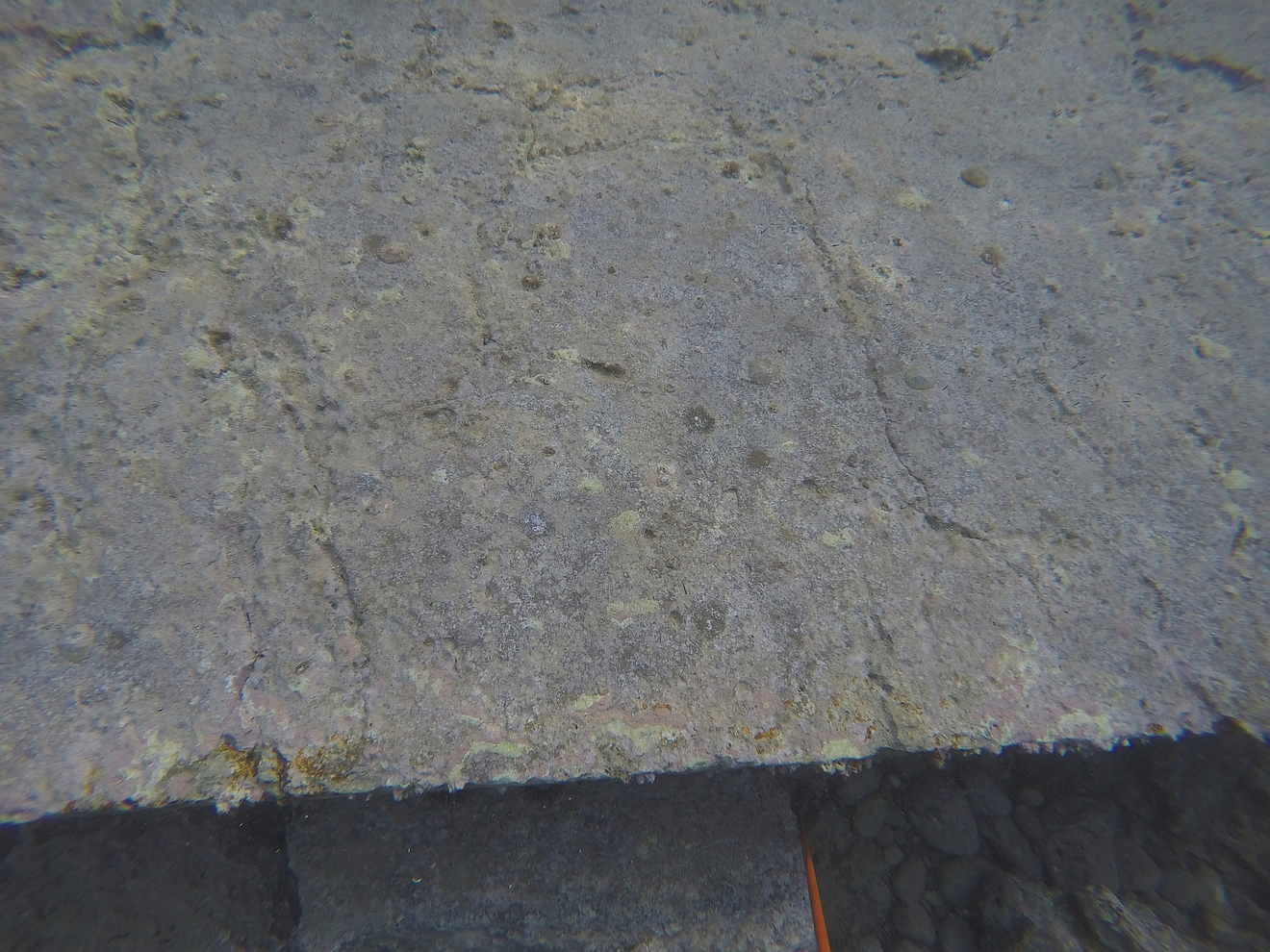}&\includegraphics[width=1.6in]{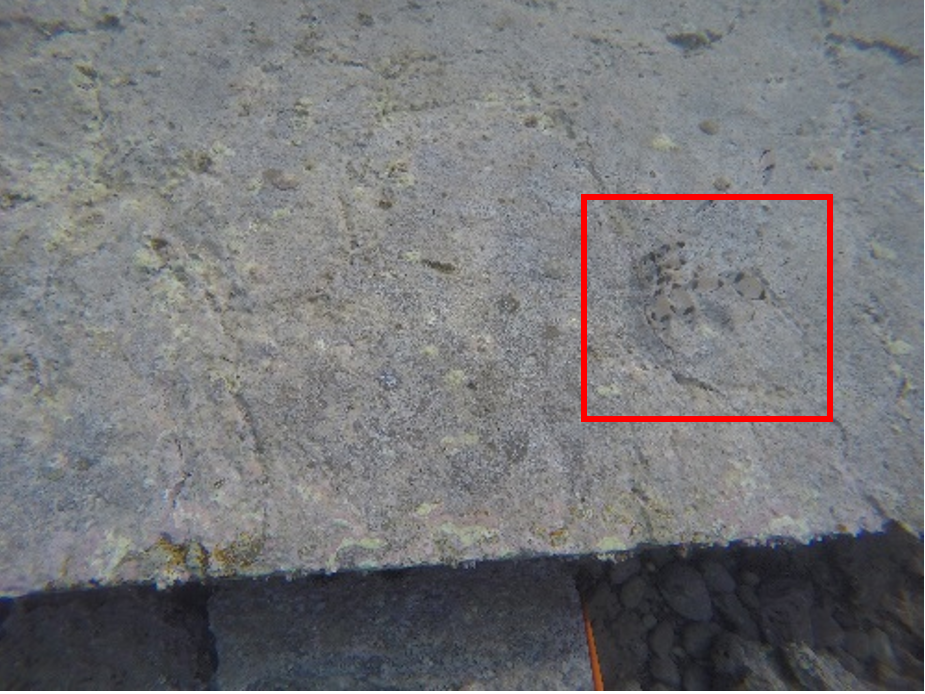}\\
(a)&(b)
\end{tabular}
\caption{The reference image created using the artificial shadow in (a) and the reference image created using the pixels having the lowest values compared with the pixels of the same position in the rest of the imagery in (b). In the red rectangle the multiple fished appearing in the reference image are highlighted.}
\label{fig:figure7.2}
\end{figure}

To capture the reference images of the dataset (Figure \ref{fig:figure7.1}, second column), artificial shadow was introduced in the scene, above the water, in order to protect the water surface from the sun rays (Figure \ref{fig:figure7.2}a). Since the images of the same branch of the dataset (i.e. left or right) are captured by the same camera position, an alternative approach was also implemented for the reference images generation. This was to create an image using the pixels having the lowest values compared with the pixels of the same position in the rest of the imagery. To perform that, images were transferred to the lab color space in order to choose the pixels with the lower luminosity. Although this approach is generating images with great quality, in some cases, in the created reference image objects appear that are apparent in one of the scenes, having lower values in the lab space compared with the rest of the pixels of this position. This led in many cases to the appearance of multiple fishes in the reference image, even in the branch of the dataset there is only one fish moving between the image instances (Figure \ref{fig:figure7.2}b). Having captured the reference images for each part of the dataset, the per-element differences $\mathrm{\Delta}(x,y)$ between the reference and each one image of the dataset containing caustics is calculated and saved in a new image, the difference image (Figure \ref{fig:figure7.1}, third column). To generate a more accurate difference image, containing only caustics and not other differences in the pixels' colors, a color transferring approach between the images containing caustics and the reference images is performed, as described in \ref{subsubsection:subsubsection6.1.2.1}. Color transferring is performed from the images with caustics to the reference image. Also, since the difference image contains noise due to scattering and passing particles, the image is smoothed with a spatial Gaussian pre-filter with a kernel size of 3 x 3 to 7 x 7 pixels, depending on the scene. To compensate changes in illumination conditions during the acquisition phase and prepare the imagery for the ground truth image generation, the difference images' pixel values are scaled and shifted so their minimum value is 0 and the maximum 255. Finally, the thresholded difference images are created (Figure \ref{fig:figure7.1}, fourth column). In order to ensure that the ground truth images do not include any false positives, a Canny edge detector \cite{185} is employed in order to facilitate the projection of the detected contours on the original images containing caustics for visual inspection (Figure \ref{fig:figure7.1}, fifth column).

Although the above procedure is very consistent for generating the ground truth images, it must be noticed that caustics are very complex physical phenomena and by their nature they have not a very specific boundary. When the camera to object distance is quite small (i.e. 0.5m) and the incidence angle of the light rays to the surface of the water is also large, their boundary is characterized by very intense chromatic aberration effects. On the contrary, when the incidence angle is closer to the perpendicular, their boundary is of the same color values as the core of the caustics -bright white. The chromatic aberration intensity is also depending on the direction of the ripples compared with the lighting rays' direction. As the angle of the ripples with the lighting direction increases, the aberration increases too. 

\subsubsection{Underwater camera calibration}
Since the collected dataset enables also the 3D reconstruction of the imaged areas of the seabed, underwater camera self-calibration \cite{198} was performed in order to deliver the retrieved parameters together with the dataset. Self-calibration techniques do not use any calibration object. Just by moving a camera in a static scene, the rigidity of the scene provides in general two constraints \cite{200} on the cameras' internal parameters from one camera displacement by using image information alone. Therefore, if images are taken by the same camera with fixed internal parameters, correspondences between three images are sufficient to recover both the internal and external parameters \cite{201}. Results are included in the dataset as a separate file. The camera models used are of OpenCV \cite{197}. For more information on the models, readers may refer to \cite{197, 198, 199}.

\section{Proposed method for pixelwise image radiometric correction}
\label{section:section6.1}
\begin{figure*}[!t]
\begin{center}
\includegraphics[width=6.5in]{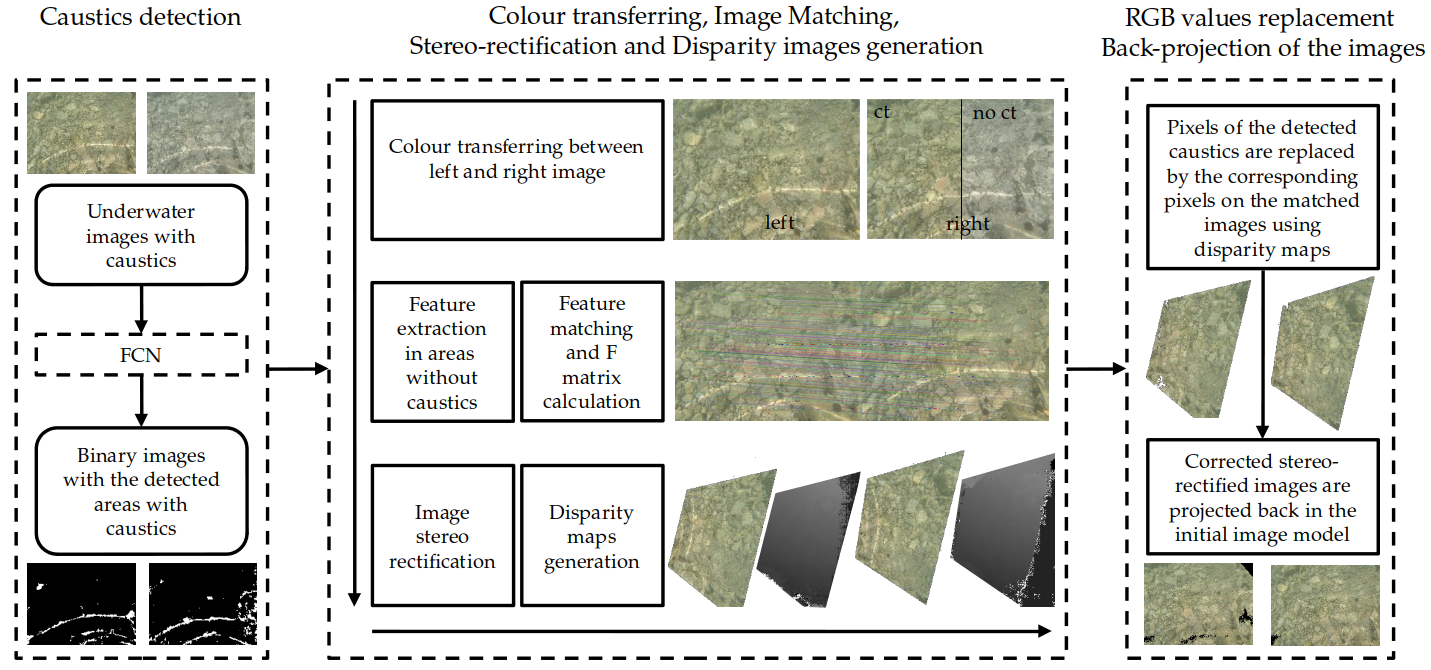}\\
\caption{The proposed method for the pixelwise correction of the rippling caustics on the underwater imagery.}
\label{fig:figure6.1}
\end{center}
\end{figure*}

To deal with the feature detection and matching problem in images with caustics, the very accurate detection of the not affected areas of the images is proposed in this work, enabling feature detection only on these areas. For the approach presented here, it is of utmost importance to maintain as much of the original RGB information of the images as possible. Next, in order to correct the affected imagery, a pixelwise method based on the stereo or the multi-view geometry is proposed. This method works on the overlapping area of the imagery and takes advantage of the pixel correspondences in this area. The overall workflow of the proposed method is illustrated in Figure \ref{fig:figure6.1} while the different modules of the method are described in detail in the following sections.

Initially, in order to detect the unaffected areas with very high accuracy, reliability and repeatability over different caustics patterns, different types of seabed, luminosity and visibility conditions, a Fully Convolutional Neural Network (FCN) classifier is exploited. The classifier is being trained using the R-CAUSTIC dataset. The selection of a specific state-of-the-art FCN as well as over shallower learning architectures is justified in Section \ref{section:section7.2}. 

Having detected the unaffected and consequently the affected areas on the initial imagery since it is a binary classification problem, a color transferring approach is performed between the consecutive or overlapping images. These images are then matched using only the key points detected on the not-affected areas of the images classified as "non-caustics", are stereo-rectified and their respective disparity maps are generated. In this step, the binary images resulted by the FCN are also re-projected using the homography matrices computed for the initial imagery rectification. Having all the necessary data available, the pixels classified as "non-caustics" are not processed while the rest of the pixels are replaced by their corresponding pixels on the matched images using the disparity maps. Finally, the corrected stereo-rectified images are projected back onto the initial camera model in order to facilitate further SfM-MVS processing and texturing. 

\subsection{Pixel classification}
Although the unaffected areas are appearing on the images as pixels having lower exposure than the caustics' pixels, for the method presented here, a sophisticated approach was chosen to be implemented. This was decided because of the very intense effects of chromatic aberration appearing on the boundaries of caustics, not being able to be detected by a simpler method such as thresholding. Moreover, it was also important to classify as not-affected areas white colored areas and reflective materials of the seabed such as e.g. marbles and GCP markers placed for photogrammetric campaigns in the shallower waters (see Section \ref{section:section7.2}). 

Since underwater imagery can be captured by very different distances from the seabed, different imaging sensors, different luminosity and visibility conditions and caustics size might vary depending on the water surface state, it was considered important for the deep fully convolutional network to be able to deal with multi-modal and multi-scale image data for semantic labeling. 

To that direction, several state-of-the-art architectures were implemented and used for training and testing, providing insights for their performance over the specific problem of caustics. As such, a similar approach to the one presented in \cite{193} which is based on SegNet \cite{192}, a U-Net \cite{225}, two FCN models with ResNet50 and ResNet101 backbones \cite{226} respectively and two DeepLabV3 \cite{227} models with a ResNet50 and ResNet101 backbones respectively were trained, evaluated and tested using the R-CAUSTIC dataset. For more information on the metrics and results see Section \ref{section:section7.2}.

\subsection{Stereo-rectification and Disparity maps generation}
\label{subsubsection:subsubsection6.1.2.3}
The most important steps of the proposed method are the accurate matching and stereo-rectification of the imagery and consequently the accurate disparity maps generation and filtering. However, prior to these steps, a color transferring approach is performed between the images of the stereopair or even the whole block.

\subsubsection{Color transferring in the CIE-$l\alpha\beta$ color space}
\label{subsubsection:subsubsection6.1.2.1}
Color transferring is an important step of the proposed method. It will prevent the appearance of visible patches of pixels in the corrected imagery and will deliver seamless corrected imagery. Indeed, due to different exposure and white color balancing of the images, it is very likely that the replaced regions are clearly recognizable by color differences creating edges in the image which are not part of the object itself. The main goal of this step is to obtain a visual consistent and pleasing corrected image, while fine details are preserved.

To apply this color transferring approach, the well-established methodology presented in \cite{189} is exploited. It is based on the $l\alpha\beta$ color spaced developed by \cite{190} which minimizes the correlation between the three image channels. This facilitates different operations in different color channels with the confidence that undesirable cross-channel artifacts will not occur \cite{189}. Additionally, this color space is logarithmic, which means to a first approximation that uniform changes in channel intensity tend to be equally detectable \cite{189}. In the implemented method, firstly, the RGB images are converted to $l\alpha\beta$ color space. There, the l axis represents an achromatic channel, while the $\alpha$ and $\beta$ channels are chromatic yellow–blue and red–green opponent channels. To achieve the color transferring, the mean and standard deviations along each of the three axes are computed for both the source and target images. These mean values are subtracted from the image data points \cite{189} i.e. the values of each channel. Consequently, the resulted data points are scaled comprising the new image by factors determined by the respective standard deviations. After this transformation, the new data points (of the new image) have standard deviations that conform to the image. Next, instead of adding the averages that were previously subtracted, the averages computed for the image are added. Finally, the result is converted back to RGB via log-LMS, LMS, and XYZ color spaces and the new image is ready for further processing.

\subsubsection{Feature detection, matching and epipolar geometry retrieval}
Next, feature detection is performed using BRISK \cite{186} detector. BRISK is adaptive, offering high quality performance as in other state-of-the-art algorithms, albeit at a dramatically lower computational cost (an order of magnitude faster than SURF in some cases) \cite{186}. To detect matchable features, the generated binary images from the previous step are exploited here. To that direction, for detecting the necessary features, BRISK \cite{186} is working only in areas that are classified as "non-caustics" in the binary mask. Even for this specific approach BRISK detector was chosen due to its lower computational cost. SIFT, SURF, ORB or other state-of-the-art detectors can also be used to detect the necessary features. 

The detected features on an image are then matched to the corresponding features on the overlapping images and the mapping of these features between these images is stored in a vector. This matching is based on n-space Euclidean distance and it is performed both from left-to-right and right-to-left for redundancy. However, since in feature matching, several blunders might occur, the RANSAC \cite{148} algorithm is utilized to identify the inliers of the obtained point correspondences. The algorithm takes all the matched points as input, formulates a mathematical model which incorporates the majority of the points, and filters out the remaining points which are considered as outliers. To accomplish that, the fundamental matrix is computed and the measure for thresholding inliers points is the distance from the epipolar line. At the end of this step, a set of matched points is found in the given scenes.

\subsubsection{Stereo-rectification and disparity maps generation}
Knowing the epipolar geometry of the overlapping images, the initial imagery is then projected to form the stereo-rectified imagery in pairs. By using exactly the same matrices which describe the epipolar geometry of the two images, the predicted binary images resulted by the FCN are also projected in their stereo-rectified form. These projections are necessary to reduce the complexity of the problem across horizontal epipolar lines and facilitate the pixelwise image radiometric correction. Since for the vast majority of the diver acquired underwater imagery, the optical axes of the cameras are not parallel to each other, in order to have horizontal epipolar lines parallel to the baseline, the reprojection of both image planes onto a common plane parallel to the baseline needs to be performed. The rectification approach followed was firstly proposed in \cite{191} and involves the decomposition of each rectifying homography into a projective and an affine component. Then the projective component which minimizes a well-defined projective distortion criterion is found. The affine component of each homography is further decomposed into a pair of simpler transforms; one designed to satisfy the constraints for rectification, the other is used to further reduce the distortion introduced by the projective component.

The stereo-rectified imagery is then used for the disparity maps generation. This is achieved by stereo-processing the imagery by Semi-Global Matching \cite{187, 188}, evaluating in 8 line directions instead of 16, to reduce the processing time. The exact approach followed is described in \cite{187} where the Consistent Semi-Global Matching (CSGM) is firstly presented. Compared with SGM, CSGM uses the same steps but also the intensity consistent disparity selection. It uses a pixelwise, mutual information based matching cost for compensating radiometric differences of input images, a feature necessary when processing images with caustics. Pixelwise matching is supported by a smoothness constraint that is usually expressed as a global cost function. CSGM performs a fast approximation by pathwise optimizations from all the 8 directions. Additionally, postprocessing steps for removing outliers, recovering from specific problems caused due to the caustics' effects and the interpolation of gaps are also applied.

Occlusions and mismatches can be distinguished as part of the left/right consistency check. Regarding the interpolation for the gaps caused by the mismatched pixel areas on the caustics, it is performed by propagating valid disparities through neighboring invalid disparity areas. To achieve the best possible interpolation, this is done similarly to SGM along paths from 8 directions. According to \cite{187}, where the implementing postprocessing was first presented, this approach emphasizes the use of all information without a preference to foreground or background. Moreover, instead of the mean, the median is used for maintaining discontinuities in cases where the mismatched area is at an object border. The implemented interpolation method has the advantage that it is independent of the used stereo matching method. Finally, median filtering can be useful for removing remaining irregularities and additionally smooths the resulting disparity image.

\subsection{RGB values replacement and back-projection of the images}
Considering that for each stereo-rectified image, the binary image of "non-caustics" and "caustics" classes and the disparity maps are available in the same projection, the pixelwise image correction can be performed. To that direction, for each pixel of an image that is not classified as "non-caustics" in the binary image, the RGB values of the corresponding pixel in the overlapping image are found, using the disparity map and vice versa. The relation between the source and the target pixels is expressed in Equation \ref{eq:equation6.1} and Equation \ref{eq:equation6.2}.

\begin{equation}
x_{target}=x_{source}-{disparity}_{source}
\label{eq:equation6.1}
\end{equation}
\begin{equation}
x_{source}={disparity}_{target}+x_{target}
\label{eq:equation6.2}
\end{equation}

where $x_{target}$ is the horizontal coordinate of the pixel in the n+1 image of the stereo pair and $x_{source}$ the horizontal coordinate of the pixel in the n image of the block. This process is performed only if the target pixel is classified as "non-caustics".

That way, the areas that are not classified as "non-caustics", are pixel-wisely corrected by exploiting the rigorous geometry of the stereo-pair. By using this approach, the disparity (x-parallax) of each pixel is taken into account, facilitating a more accurate and reliable correction approach that is not adversely affecting the later SfM and MVS steps, but it improves them. This is achieved as this horizontal displacement between rectified feature points is related to the depth of the feature. This way the method can be used to recover the images over 3D structures without affecting the 3D position of the SfM and MVS processing calculated subsequent.

\section{Experimental Evaluation}
\label{section:section7.2}
In this section, the proposed method for pixel-wise image correction is tested end evaluated over the R-CAUSTIC dataset and images collected from other random constructions, proving its robustness, accuracy and reliability.

\subsection{Pixel classification}
To form a reliable image pixel classification model, the dataset presented in Section \ref{section:section7.1} was used to train, cross-validate and test SegNet \cite{192}, U-Net \cite{225}, two FCNs with ResNet50 and ResNet101 backbones \cite{226} respectively and two DeepLabV3 \cite{227} models with a ResNet50 and ResNet101 backbones respectively. To that direction and in order to prove and evaluate their generalization potential over different types of scenes, seven training-testing approaches were performed for each of the FCNs. To achieve that, the subsets were categorized based on the similarity of the scene; subsets 1, 2 and 3 are considered to have similar background, although caustics are of very different scale and complexity, subsets 5 and 7 are also considered similar. The rest of the subsets are not similar to each other. They represent totally different types of seabed and caustics.

\subsubsection{Training and cross-validating the models}
Considering the above, for the first case, the models are trained over subsets 4, 5, 6 and 7 and testing is performed over subsets 1, 2 and 3. For the second case, the models are trained over subsets 3, 4, 5, 6 and 7 and testing is performed over subsets 1 and 2. The third case involves training on 1, 2, 3, 4, and 6 and testing on subsets 5 and 7. In the fourth approach, the models are trained over subsets 1, 2, 3, 5, and 7 and tested on subsets 4 and 6. Finally, in the fifth, sixth and seventh approach, the models are trained over only subsets 1, 3 and 6 respectively and tested over the same subsets; 2, 4, 5 and 7. The number of the images used for training these three models (three for each network) are exactly the same with the previous training approaches. These last three approaches aimed to prove the necessity of the variability of the scenes of the dataset and to reply to the question whether it is better to train the model using $N$ images coming from one subset or using $N$ images coming from many subsets.

For training the FCNs, a sliding window approach to extract 128 x 128 patches was used. The stride of the sliding window defines the size of the overlapping regions between two consecutive patches. At training time, a smaller stride allows us to extract more training samples and acts as data augmentation. At testing time, a smaller stride allows us to average predictions on the overlapping regions, which reduces border effects and improves the overall accuracy. During training, cross-validation and testing, a 32 pixels stride was used. Models are implemented using the PyTorch framework. Torch is a scientific computing framework with wide support for machine learning algorithms that puts GPUs first. All the models were trained over 30 epochs using Stochastic Gradient Descent (SGD) with a base learning rate of 0.01, a momentum of 0.9, a weight decay of 0.0005 and a batch size of 60. The learning rate was divided by 10 after 20 epochs. Regrading the number of the epochs of training, various applied approaches indicated that more than 30 epochs offer nothing more to the network's performance. To this direction, the network was trained on 30 epochs where for each epoch were used 10.000 samples acquired from 10 images of each subset over 167 iterations. For the training approaches on which only one subset was used, the same number of samples was retrieved from 50 images. 

\begin{figure}[!h]
\centering
\includegraphics[width=3.5in]{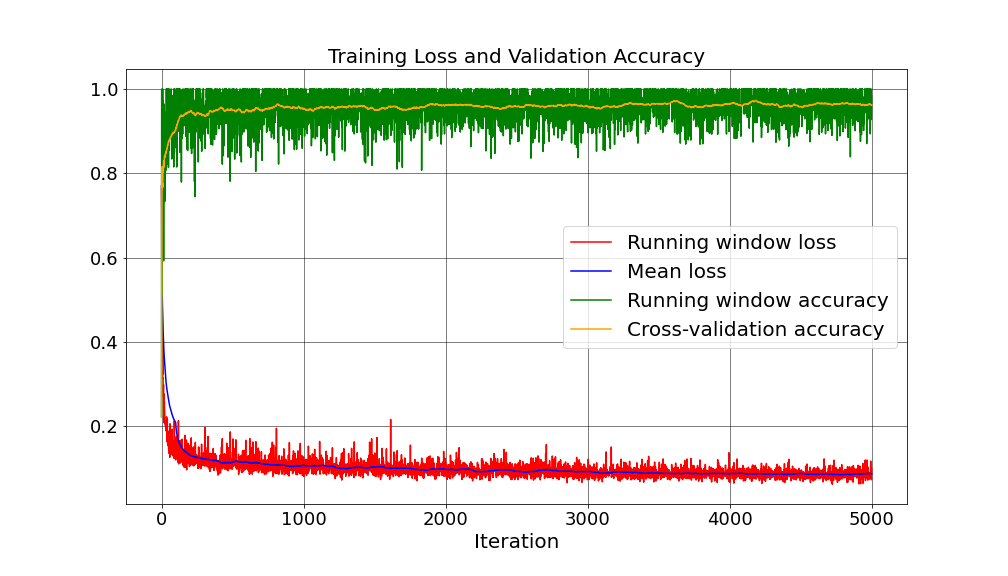}
\caption{Training loss and cross-validation accuracy for training the FCN with the ResNet101 backbone over subset 1.}
\label{fig:figure7.12}
\end{figure}

Figure \ref{fig:figure7.12} presents the training loss and cross-validation accuracy for training FCN-ResNet101 on subset 1. Lines in green color represent the running-window cross-validation accuracy while lines in orange the cross-validation accuracy on all the validation data for each 100 iterations. Lines in red represent the training loss for the same running-window while lines in blue represent the mean loss for each iteration. In Figure \ref{fig:figure7.12}, which is representative for the majority of the performed training approaches, one can notice that the cross-validation accuracy (orange line) is over 92-94\% and the mean training loss (blue line) is less than 0.15 even from the first 2500 iterations (15 epochs). However, moderate oscillation in the running window cross-validation accuracy and training loss is present in most of the multi-subset training approaches. 

The oscillation observed in the same figure is directly related to the complexity and the clarity of the caustics on the subsets used for training. To investigate the source of this oscillation, initially the base learning rate is altered. By the tests performed it is concluded that the reduction of the learning rate did not decreased the oscillation of the accuracy but on the contrary it increased it while the increase of the learning rate did not affect it. This implies that the oscillation of the running-window values is caused by the variance of the cross-validation data and especially the use of a random sample. Increasing the window size or reducing batch size will effect in reduction of the oscillation, however this is not the case for the performed experiments. Regarding the accuracy of all the cross-validation performed every 100 iterations for all the validation data and the mean loss, no remarkable oscillation is noticed, proving the reliability of the trained models.

\subsubsection{Testing the models}
To test the different models trained over unseen subsets of the dataset, two metrics are used; \textit{F1 score} which is a single metric that combines \textit{recall} and \textit{precision} using the harmonic mean and \textit{accuracy} which is the ratio of the correctly labeled subjects to the whole pool of subjects.

Table \ref{table:table7.1} presents the various metrics calculated after testing the several trained models over 30 epochs. It is obvious that for all the trained models, the achieved metrics for "non-caustics" pixel classification are quite high, indicating their great generalization. 

\begin{table*}[!h]
\caption{Testing performance after 30 epochs of training.}
\label{table:table7.1}
\centering
\renewcommand{\arraystretch}{0.8}
\begin{tabular}{@{}|c|c|c|c|c|c|@{}}
\hline
  {\begin{tabular}[c]{@{}c@{}}Training subsets\end{tabular}} &
  {\begin{tabular}[c]{@{}c@{}}Testing subsets\end{tabular}} &
  {\begin{tabular}[c]{@{}c@{}}Method  (\%)\end{tabular}} &
  {\begin{tabular}[c]{@{}c@{}}Caustics F1   (\%)\end{tabular}} &
  {\begin{tabular}[c]{@{}c@{}}Non-Caustics F1 (\%)\end{tabular}} &
  {\begin{tabular}[c]{@{}c@{}}Total Accuracy (\%)\end{tabular}} \\ \hline
4,5,6,7   & 1,2,3       & SegNet & 77.64 & 96.46 & 93.89 \\
4,5,6,7   & 1,2,3       & UNet & 76.53 & 96.32 & 93.64 \\
4,5,6,7   & 1,2,3       & FCN-ResNet50 & 78.61 & 96.48 & 93.96 \\
4,5,6,7   & 1,2,3       & FCN-ResNet101 & \textbf{79.16} & \textbf{96.50} & \textbf{94.00} \\
4,5,6,7   & 1,2,3       & DeepLabV3+ResNet50 & 78.62 & 95.79 & 92.96 \\
4,5,6,7   & 1,2,3       & DeepLabV3+ResNet101 & 78.98 & 95.98 & 93.25 \\\hline
3,4,5,6,7 & 1,2         & SegNet & \textbf{83.59} & \textbf{97.20} & \textbf{95.22} \\
3,4,5,6,7 & 1,2         & UNet & 81.88 & 97.02 & 94.88 \\
3,4,5,6,7 & 1,2         & FCN-ResNet50 & 83.22 & 96.91 & 94.82 \\
3,4,5,6,7 & 1,2         & FCN-ResNet101 & 83.39 & 97.13 & 95.10 \\
3,4,5,6,7 & 1,2         & DeepLabV3+ResNet50 & 81.25 & 96.30 & 93.82 \\
3,4,5,6,7 & 1,2         & DeepLabV3+ResNet101 & 82.94 & 96.83 & 94.65 \\\hline
1,2,3,4,6 & 5,7         & SegNet & 78.45 & 97.44 & 95.42 \\
1,2,3,4,6 & 5,7         & UNet & 79.08 & 97.44 & 95.44 \\
1,2,3,4,6 & 5,7         & FCN-ResNet50 & 79.72 & 97.47 & 95.50 \\
1,2,3,4,6 & 5,7         & FCN-ResNet101 & \textbf{80.56} & \textbf{97.61} & \textbf{95.74} \\
1,2,3,4,6 & 5,7         & DeepLabV3+ResNet50 & 79.72 & 9.25 & 95.16 \\
1,2,3,4,6 & 5,7         & DeepLabV3+ResNet101 & 80.18 & 97.31 & 95.27\\\hline
1,2,3,5,7 & 4,6         & SegNet & 73.03 & 96.58 & 93.93 \\
1,2,3,5,7 & 4,6         & UNet & 73.19 & 96.57 & 93.92 \\
1,2,3,5,7 & 4,6         & FCN-ResNet50 & 77.45 & 97.05 & 94.78 \\
1,2,3,5,7 & 4,6         & FCN-ResNet101 & 77.87 & \textbf{97.06} & \textbf{94.81} \\
1,2,3,5,7 & 4,6         & DeepLabV3+ResNet50 & 77.97 & 96.77 & 94.36 \\
1,2,3,5,7 & 4,6         & DeepLabV3+ResNet101 & \textbf{78.19} & 96.80 & 94.42 \\\hline
1         & 2,4,5,7 & SegNet & \textbf{72.99} & \textbf{96.30} & \textbf{93.49} \\
1         & 2,4,5,7 & UNet & 73.26 & 96.27 & 93.45 \\
1         & 2,4,5,7 & FCN-ResNet50 & 71.64 & 95.84 & 92.74 \\
1         & 2,4,5,7 & FCN-ResNet101 & 72.20 & 95.97 & 92.96 \\
1         & 2,4,5,7 & DeepLabV3+ResNet50 & 70.91 & 95.56 & 92.30 \\
1         & 2,4,5,7 & DeepLabV3+ResNet101 & 71.57 & 95.74 & 92.58 \\\hline
3         & 2,4,5,7 & SegNet & 68.51 & 95.08 & 91.49 \\
3         & 2,4,5,7 & UNet & 67.62 & 94.90 & 91.19 \\
3         & 2,4,5,7 & FCN-ResNet50 & 68.60 & 95.10 & 91.52 \\
3         & 2,4,5,7 & FCN-ResNet101 & \textbf{69.17} & \textbf{95.22} & \textbf{91.72} \\
3         & 2,4,5,7 & DeepLabV3+ResNet50 & 65.42 & 93.97 & 89.74 \\
3         & 2,4,5,7 & DeepLabV3+ResNet101 & 66.72 & 94.35 & 90.35 \\\hline
6         & 2,4,5,7 & SegNet & 67.55 & 96.45 & 93.61 \\
6         & 2,4,5,7 & UNet & 63.01 & 96.22 & 93.14 \\
6         & 2,4,5,7 & FCN-ResNet50 & 66.52 & 96.56 & 93.77 \\
6         & 2,4,5,7 & FCN-ResNet101 & 64.71 & 96.58 & 93.77 \\
6         & 2,4,5,7 & DeepLabV3+ResNet50 & 69.84 & 96.42 & 93.60 \\
6         & 2,4,5,7 & DeepLabV3+ResNet101 & \textbf{70.81} & \textbf{96.84} & \textbf{94.30} \\\hline
\end{tabular}
\end{table*}

It is also evident that the models trained only on subsets 3 or 6 are achieving lower metrics compared to the rest, proving the necessity of the specific dataset as a whole. It is reminded here that the main goal here was the very accurate detection of the not affected areas of the images, enabling feature detection and color values transferring only from those areas.

Overall, the results delivered by all the trained models are very satisfying, classifying with very high accuracy and reliability the pixels of the underwater images. Differences on the sensitivity of the models are expected, however this highlights the need to train a final model using all the available subsets. However, the slight prevalence of the models trained on the FCN-ResNet101 should be mentioned, while SegNet follows.

The small variations observed in the metrics were expected due to the diverse characteristics of the subsets used for training and testing and the different FCN architectures used for training. It is also an indication of the performance of the trained models over different types of caustics. However, these variations are not affecting the overall accuracy of the models in a severe degree. Actually, the differences in the F1 score for Caustics and "non-caustics" are mostly attributed to the indefinite boundary of caustics due to the chromatic aberration effect, and as such the differences between the predicted and the ground truth values, especially in subsets 3 and 6. This is also the reason for the lower accuracy achieved in "caustics" class, since the chromatic aberration affects a larger percentage of this class, compared to the "non-caustics" class.

\begin{figure}[!h]
\centering
\includegraphics[width=3.5in]{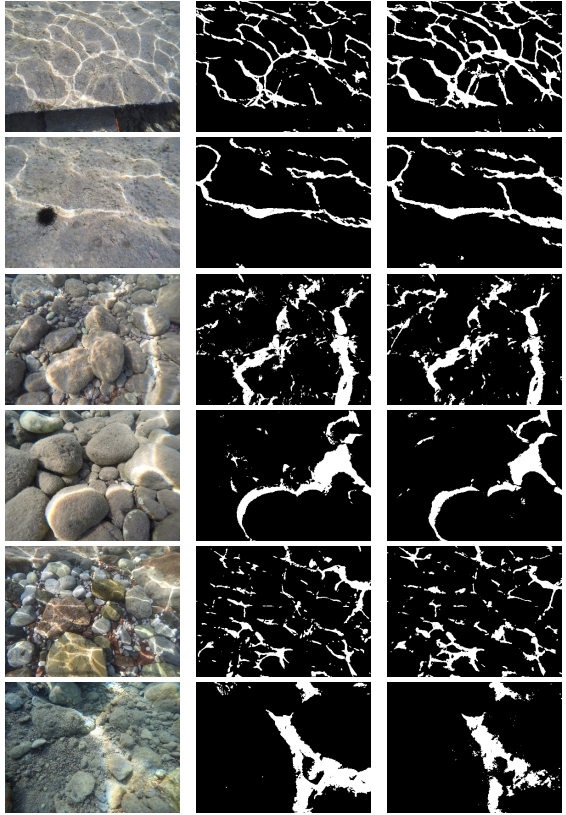}
\caption{Typical examples of the RGB images (first column), the ground truth binary images (second column) and the predicted binary images (third column) using the models already described.}
\label{fig:figure7.4}
\end{figure}

\begin{figure}[!h]
\centering
\includegraphics[width=3.5in]{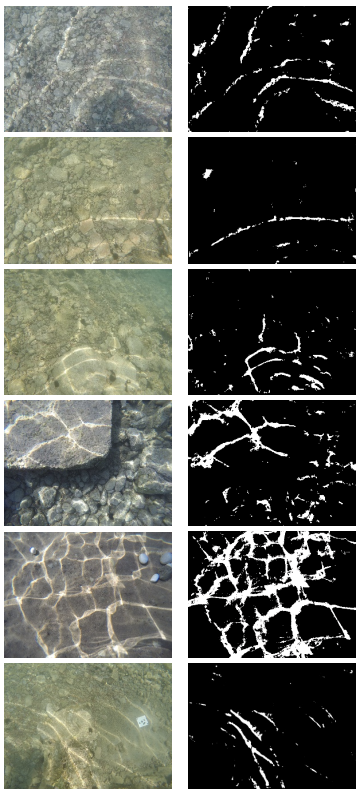}\\
\caption{Various real-world RGB images (left column) and the respective predictions using the model trained on all the available subsets (right column).}
\label{fig:figure7.5}
\end{figure}

In Figure \ref{fig:figure7.4} the input RGB images with the respective ground truth and SegNet's predictions are illustrated. For the first and the second row of the figure which is showing the predictions over an image from 1 and 2 subset respectively, the binary image is predicted by the model trained on 3, 4, 5, 6 and 7 subsets. Especially in the first row, it can be noticed that the pixels classified as "caustics" exceed those in the ground truth. These results come from the training of the model in more close-range imagery, giving it the ability to detect the intense chromatic aberration on the boundaries of the caustics. However, this cannot be achieved in the ground truth image. Following, the third and the fourth row present the predicted binary masks using the model trained over the 1, 2, 3, 4, and 6 subsets for images belonging to the fifth and the seventh subsets respectively. Finally, the fifth and the sixth rows are illustrating the predicted binary masks using the model trained on the 1, 2, 3, 5 and 7 subsets for images belonging to the fourth and the sixth subsets respectively.

Additionally, an FCN-ResNet101 model was trained exploiting all the available subsets, in order to be used for classifying unseen underwater imagery. For training this model, the same parameters as before are used. Results over images captured for real world underwater photogrammetric applications in shallow waters are presented in Figure \ref{fig:figure7.5}. There, it is obvious that the trained model can generalize over different types of seabed and caustics with high reliability. A very important outcome is also that the model achieves to classify really bright areas of the image that are not caustics as "non-caustics" correctly. This can be seen on the last two images of Figure \ref{fig:figure7.5} where the white stones on the seabed and a white ground control point (GCP) are classified as "non-caustics". This example with the GCP is of really high importance for the underwater photogrammetric applications.

\subsection{Deep over shallow ML architectures}
Concerning the selection of a deep architecture for addressing the image classification problem, before the decision that an FCN would be more appropriate for the proposed solution, several tests were performed using shallower ML architectures. Training and testing was performed for classifying the underwater imagery using AdaBoost \cite{202} and specifically AdaBoost-SAMME \cite{207} having Decision Trees \cite{203} as the base estimator, Decision Trees \cite{203}, K-Nearest Neighbors (KNN) \cite{204}, Quadratic Discriminant Analysis (QDA), Random Forests \cite{205} and linear Support Vector Machines \cite{206}. Also, a simple thresholding approach was performed. Regarding the thresholding, its value was changed over the different images while for the rest of the methods, the default values as set in \cite{88} were used. Resulting metrics from training over subsets 4, 5, 6 and 7 and testing over subsets 1, 2 and 3 are presented in Table \ref{table:table7.2}.

\begin{table}[!h]
\caption{The testing performance of the various methods. Number in bold are the best metrics achieved. F1 stands for "non-caustics" F1 score}
\label{table:table7.2}
\centering
\begin{tabular}{@{}|c|c|c|@{}}
\hline
  {\begin{tabular}[c]{@{}c@{}}Method\end{tabular}} &

  {\begin{tabular}[c]{@{}c@{}}F1 (\%)\end{tabular}} &
  {\begin{tabular}[c]{@{}c@{}}Accuracy (\%)\end{tabular}} \\ \hline
AdaBoost       & 89.02 & 80.48 \\
Decision Trees & 87.37 & 77.93 \\
KNN            & 87.61 & 78.30 \\
QDA            & 84.34 & 73.49 \\
Random Forests & 87.56 & 78.21 \\
Linear SVM     & 89.61 & 81.40 \\
{SegNet}       & 96.46    & 93.89         \\
{U-Net}        & 96.32       & 93.64        \\
{FCN+ResNet50} & 96.48&   93.96     \\
{FCN+ResNet101} & 96.50      &   94.00     \\
{DeepLabV3+ResNet50}&  95.79          &    92.96     \\
{DeepLabV3+ResNet101}&  95.98         &    92.96     \\\hline
\end{tabular}
\end{table}

As can be seen in Table \ref{table:table7.2}, the metrics achieved by all the FCNs outperform the rest of the methods used for image classification. The Linear SVM follows with almost 10\% less accuracy and after that, AdaBoost comes with almost 13\% less accuracy. By comparing the precision score of the rest of the classifiers, they are achieving quite similar percentages, however this is not the case for the recall and the F1 scores where only the linear SVM and the AdaBoost classifiers are quite close. It is also evident that AdaBoost is outperforming the results of the Decision Trees, however this was expected.

\begin{figure}[!h]
\centering
\includegraphics[width=3.5in]{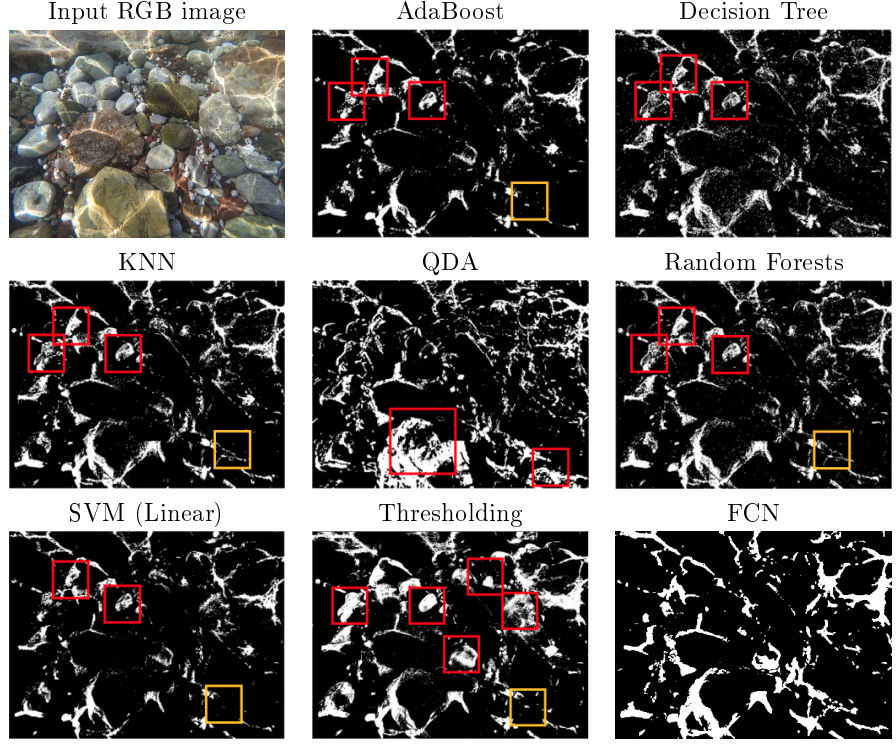}\\
\caption{The input RGB image together with the predictions of the tested architectures for an image from subset 4.}
\label{fig:figure7.6}
\end{figure}

\begin{figure}[!h]
\centering
\includegraphics[width=3.5in]{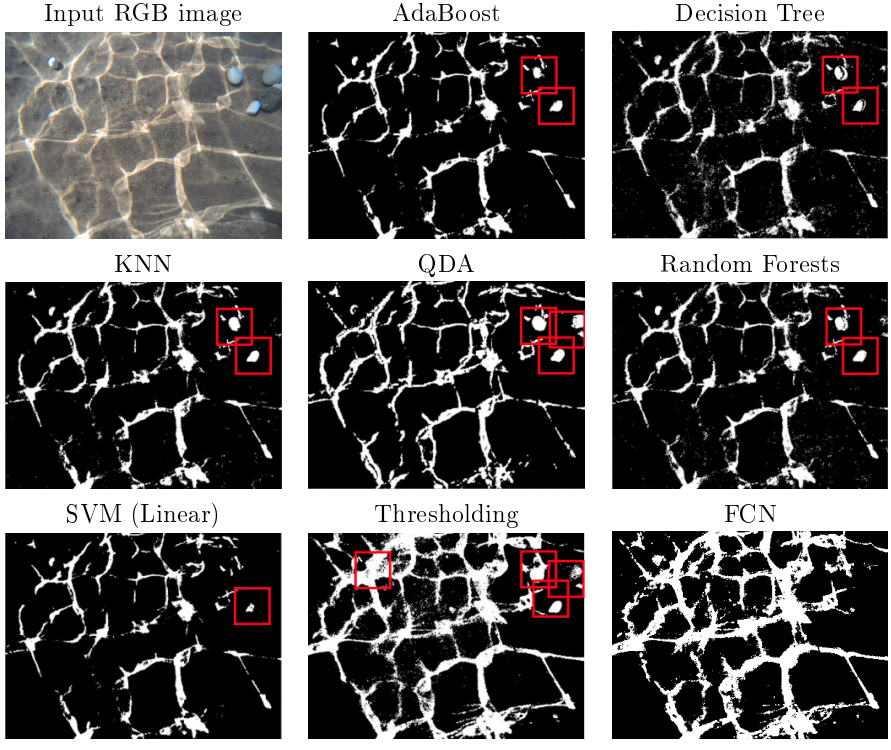}\\
\caption{The input RGB image together with the predictions of the tested architectures for an image of a real world application, not contained in the dataset.}
\label{fig:figure7.7}
\end{figure}

For the majority of the performed tests, the shallower classifiers as well as the thresholding were not able to distinguish between caustics and other bright artifacts on the scene i.e. white rocks with high accuracy. Moreover, they did not succeed in reliably detecting the boundary of the caustics when it is characterized by the intense chromatic aberration effect. This is also reflected in the achieved metrics in Table \ref{table:table7.2}. Two typical examples are presented in Figure \ref{fig:figure7.6} and Figure \ref{fig:figure7.7}.

There, the input RGB image together with the predictions of the tested shallow architectures are given. In the red rectangles some typical cases of false negatives i.e. pixels that are wrongly classified as "caustics" are highlighted. On the other hand, orange rectangles are indicating areas of false positive, i.e. pixels thar are wrongly classified as "non-caustics". This, together with the performed experiments over the different subsets, highlights the difficulty of those methods to generalize over the different types of seabed and caustics.

The metric and visual results of the above experiments, justify the use of a deeper architecture for classifying the underwater imagery affected by caustics, as also used in \cite{123}. Even if the resulted metrics of most of the tests performed using the shallower architectures are lower than those achieved by the exploited FCN architecture, they do not prohibit their use. However, one have to consider the specific goal of the application. As such, for a more close to real time application, where only key point detection and matching is needed i.e. visual odometry etc., AdaBoost, Decision Trees, KNNs, QDA or even simple thresholding could be used due to their minimal prediction times, compared to the SVMs and the FCN architectures.

However, when it comes to the maximum improvements on the sparse and the dense 3D reconstruction using the less necessary images, a deep architecture seems to be the best solution. By classifying the pixels of the underwater imagery with the highest accuracy, only the pixels that it is really necessary to be replaced are undergoing the correction process, keeping as much as possible of the original imagery unprocessed. This way, there is more area on the image available for the keypoint detectors, ensuring a more robust SfM process. Additionally, as already proved, the FCN, achieves to detect in an accurate and reliable degree the intense chromatic aberration effects on the boundaries of the caustics, something really important for facilitating a more realistic corrected imagery and as will be shown in Subsection \ref{subsection:subsection7.2.4}, enabling the generation of a more complete dense 3D point cloud.

\subsection{Pixelwise image radiometric correction}
\begin{figure}[!t]
\centering
\includegraphics[width=3.5in]{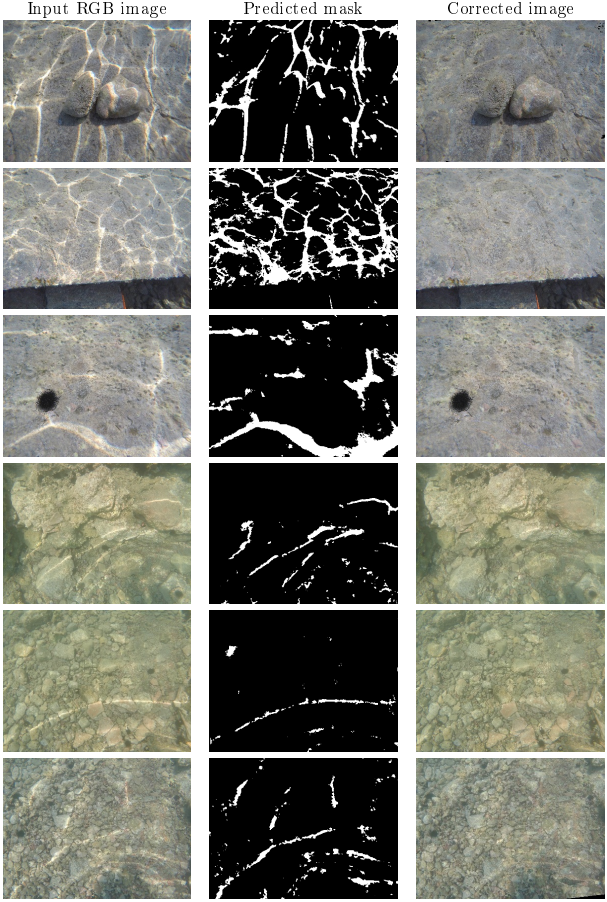}\\
\caption{Examples of corrected images. The first column depicts the original images with caustics, the second column depicts the binary images predicted by the trained FCN model and the third column demonstrates the corrected imagery.}
\label{fig:figure7.8}
\end{figure}

In this section, results regarding the pixelwise image correction are presented and evaluated. Figure \ref{fig:figure7.8} presents typical examples of corrected images.

The first column depicts the original images with caustics, the second column depicts the binary images predicted by the trained FCN model and the third column demonstrates the corrected imagery. By observing the corrected imagery, it is clear that the proposed method achieves impressive visual results, since in the vast majority of the corrected images the rippling caustics are not apparent, and the replaced pixels are not clearly obvious. However, in some cases like the one presented in the first row of Figure \ref{fig:figure7.8}, the corrected areas are obvious. This is a result of an inadequate color transferring approach, indicating great differences in luminosity between the overlapping images used for the correction. In fact, in some cases, these differences cannot be compensated by the color transferring only. However, as will be shown by the experiments presented in Subsection \ref{subsection:subsection7.2.4}, this is not affecting the 3D reconstruction performance in a measurable degree. To solve this issue, Poisson blending \cite{208} could be implemented on the boundary of the corrected with the uncorrected areas, however, this would affect the pixel values in an  uncontrollable degree, affecting even more the SfM-MVS processes.

\begin{figure}[!t]
\centering
\begin{tabular}{ccc}
\includegraphics[width=1.6in]{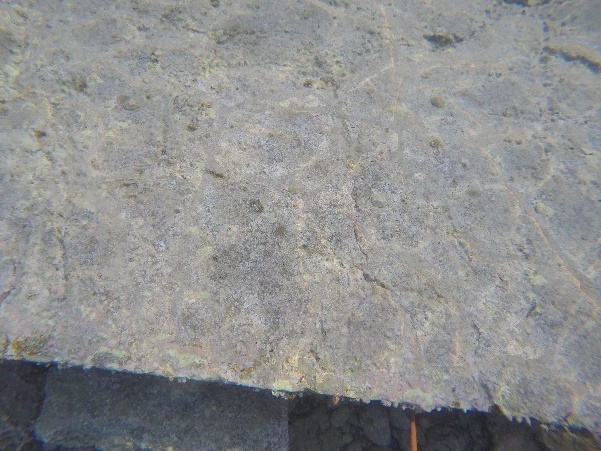}&\includegraphics[width=1.6in]{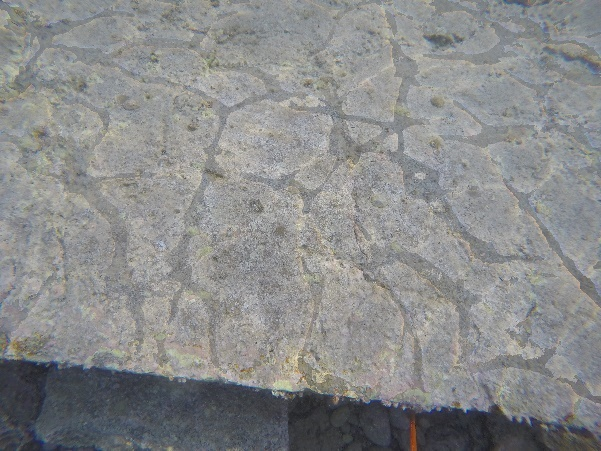}\\
{(a)}&{(b)}\\
\end{tabular}
\caption{A corrected image by the proposed methodology with the color transferring step (a) and the same image corrected by the proposed methodology without applying the color transferring step (b).}
\label{fig:figure7.9}
\end{figure}

In order to demonstrate the importance of the color transferring step proposed in this methodology, in cases where no extreme differences between the source and the target images are apparent, a typical example is given in Figure \ref{fig:figure7.9}. There, a corrected image created by the proposed methodology with the color transferring step is presented in (a) while the same image corrected by the proposed methodology without applying the color transferring step is presented in (b). It is obvious that in the case where the color transferring is not applied, the replaced pixels are obviously enough, since they are characterized by lower RGB values.

Although it is not clearly obvious in the scale at which the images are presented in Figure \ref{fig:figure7.8}, the proposed approach achieved really high accuracy in the pixel replacement process. When having a closer look to the corrected imagery, no offsets and other pixel displacements are obvious, especially in formations on the scene, being continuous between a corrected and a non-corrected area of the image.

\subsection{Improving 3D reconstruction}
\label{subsection:subsection7.2.4}
\begin{figure*}[!ht]
\centering
\includegraphics[width=5in]{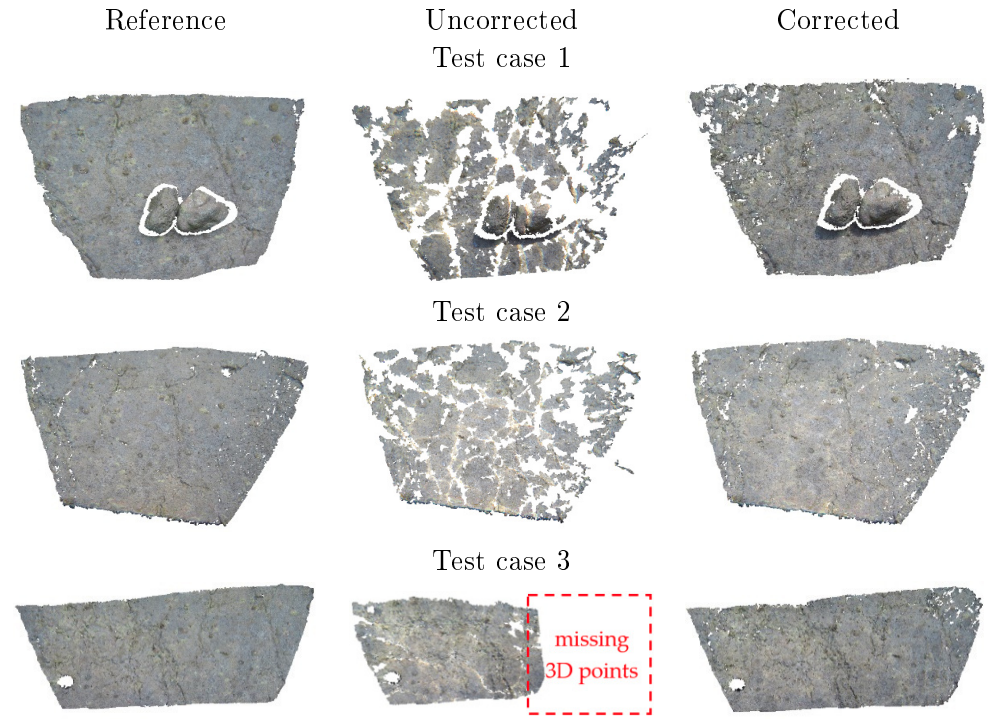}\\
\caption{The 3D reconstructions using the reference imagery (left column), the uncorrected imagery (middle column) and the corrected imagery by the proposed methodology (right column). Each row refers to a different test case.}
\label{fig:figure7.10}
\end{figure*}

\begin{figure}[!ht]
\centering
\includegraphics[width=3.5in]{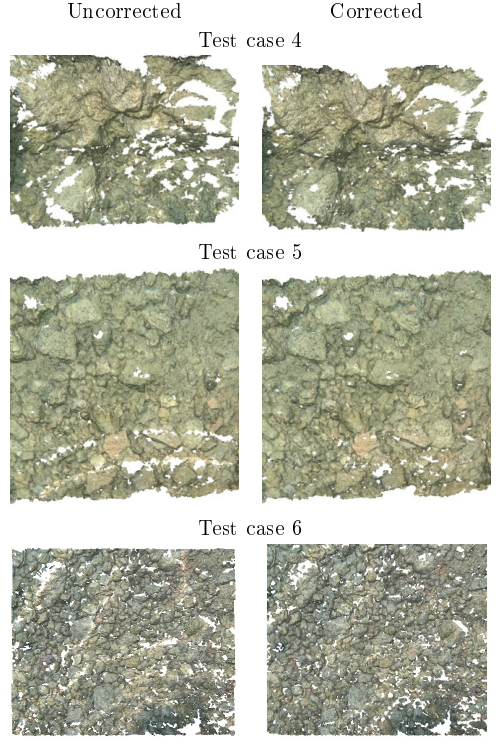}\\
\caption{The resulting 3D point clouds for the rest three tests performed over real-world imagery. In the left column the 3D point cloud generated using the original uncorrected imagery is presented while in the right column, the respective 3D point cloud of the corrected imagery is presented.}
\label{fig:figure7.11}
\end{figure}

Following the pixelwise correction of the underwater imagery, the corrected data were processed with a commercial SfM-MVS software for evaluating the improvements on the 3D reconstruction, which was the initial objective of the effort. To that direction, six different test cases were processed (Figure \ref{fig:figure7.10} and Figure \ref{fig:figure7.11}). Typical images for these six test cases can be seen in Figure \ref{fig:figure7.8}.

The first three test cases are using images of the already presented benchmark dataset; the first test case is using tri-stereo imagery from  subset 3, the second testcase is using stereo imagery from subset 1 while for the third test case multiple-view stereo imagery (7 consecutive images) from subset 2 was used. These three test cases will facilitate the evaluation of the proposed correction methodology in terms of 3D reconstruction improvement, compared also to the generated 3D point clouds using the already available reference imagery. The 3D reconstruction results are presented in Figure \ref{fig:figure7.10}. Results of the rest of the four subsets presented in Section \ref{section:section7.1} are not presented here since due to the complexity of the scene and the poor texture of the smooth and glossy rocks, the improvements are not easy to be highlighted, since the 3D point clouds of the reference imagery are incomplete too. As such, the subsets presented in Figure \ref{fig:figure7.10} are those that are generating the most complete 3D reconstructions, facilitating a detailed and direct comparison of the results.

By comparing the 3D point clouds generated using the reference imagery (left column of Figure \ref{fig:figure7.10}) to the 3D point clouds generated using the uncorrected and the corrected imagery, significant differences can be noticed. It is clear that caustics are preventing a proper 3D reconstruction of the scene, since when images affected by them are used, the resulting 3D point clouds are incomplete for all the tested cases. More specifically, Dense Image Matching (DIM) algorithms are failing to match the corresponding points of the affected areas and only the unaffected areas are appearing in the 3D point cloud. It is also important to highlight that as reported also in Table \ref{table:table7.3}, when using the uncorrected imagery, not all of the images were aligned for the test case 1 and test case 3. In the latter, this resulted in a much less covered area by the 3D point cloud.

Coming to the third column, it is obvious that when the imagery corrected by the proposed methodology is used, the completeness of the 3D point cloud is improved to a great degree, delivering point clouds very similar to the ones generated by the reference imagery. Some insignificant differences can be observed on the perimeter of the reconstructed area. These differences are not attributed to some defect of the corrected imagery but on the extremely intense chromatic aberration that is apparent at the areas of the image having large radial distance when the luminosity of the scene is increased. This effect is not that intense on the reference images since they are characterized by lower illumination. In these cases, chromatic aberration is caused by the lens of the camera, as a result of the different refractive indices of the lens for each wavelength of light.

In Figure \ref{fig:figure7.11}, three different tests performed over imagery used for real underwater 3D reconstruction tasks are presented. Again, as in the first three test cases, when using real world imagery, the imagery resulted by the proposed methodology achieved to deliver a more complete 3D point cloud in the areas covered by caustics. 

To facilitate a deeper evaluation of the improvements on the 3D reconstruction, all the 3D point clouds presented above were imported into Cloud Compare freeware \cite{81} for further investigation. In particular, the following parameters and statistics that are widely used also in the literature for evaluating 3D point clouds \cite{14, 9, 10}, were computed for each point cloud:\\

\textbf{Total number of sparse and dense points}. All the 3D points of the point cloud were considered for this metric, including any outliers and noise \cite{81}. For the purposes of the work presented here, the total number of 3D points reveals the effect of the correction methodology on the matchable pixels among the images. The more corresponding pixels are found in the DIM step on the images, the more points are generated. Higher values of total number of points are considered better in these cases; however, this should be crosschecked with the surface density metric, since it might be an indication of noise on the point cloud.

\textbf{Surface Density}. The density is estimated by counting the number of neighbours N (inside a sphere of radius R) for each point \cite{81}. The surface density used for this evaluation is defined as $\frac{N}{Pi\ \times\ R^2}$, i.e., the number of neighbours divided by the neighbourhood surface. Cloud Compare \cite{81} estimates the surface density for all the points of the cloud and then it calculates the average value for an area of 1m2 in a proportional way. Surface density is considered to be a positive metric, since it defines the number of the points on a potential generated surface, excluding the noise being present as points out of this surface. This is also the reason of using the surface density metric instead of the volume density metric.

\textbf{Roughness}. For each point, the "roughness" value is equal to the distance between this point and the best fitting plane computed on its nearest neighbour \cite{81}, which are the points within a sphere centred on the point. Roughness is considered to be a negative metric since it is an indication of noise on the point cloud, assuming an overall smooth surface.

\begin{table*}[!ht]
\centering
\caption{The evaluation metrics between the uncorrected and the corrected images. D is the surface density and R the roughness. Numbers for these two metrics are relative.}
\label{table:table7.3}
\begin{tabular}{@{}|l|c|c|c|c|c|c|c|c|c|@{}}
\hline
{{Test case}} &
  {{\begin{tabular}[c]{@{}c@{}}Images\\ aligned/\\total\end{tabular}}} &
  {{\begin{tabular}[c]{@{}c@{}}Caustics \\ in pixels\\ {[}\%{]}\end{tabular}}} &
  {{\begin{tabular}[c]{@{}c@{}}Matched\\ points\end{tabular}}} &
  {{\begin{tabular}[c]{@{}c@{}}Matched \\ points \\ change {[}\%{]}\end{tabular}}} &
  {{\begin{tabular}[c]{@{}c@{}}Dense\\ points\end{tabular}}} &
  {{\begin{tabular}[c]{@{}c@{}}Dense  \\ points \\ change \\ {[}\%{]}\end{tabular}}} &
  {{D}} &
  {{\begin{tabular}[c]{@{}c@{}}D \\ change\\ {[}\%{]}\end{tabular}}} &
  {{R}} \\
\hline
 {1}         uncorrected        & 2/3 & {28.1} & 1327 &       & 37505   &       & 7764.99   &       & 0.003 \\
                            {1 }corrected         & 3/3 &                       & 2210 & +66.5 & 72261   & +92.7 & 8140.25   & +4.87 & 0.003  \\ 
{2}          uncorrected        & 2/2 & {15}   & 606  &       & 1783027 &       & 106908.18 &       & 0.001 \\
                            {2} corrected         & 2/2 &                       & 820  & +35.3 & 2118934 & +18.8 & 106718.97 & -0.17 & 0.001 \\
{3}          uncorrected        & 5/7 & {15.4} & 2562 &       & 556926  &       & 12364.76  &       & 0.001 \\
                            {3} corrected         & 7/7 &                       & 4173 & +62.9 & 6137345 & +1102 & 78761.6   & +637  & 0.001 \\
{4}          uncorrected        & 2/2 & {8}    & 3920 &       & 188203  &       & 1590.53   &       & 0.007 \\
                            {4} corrected         & 2/2 &                       & 4916 & +25.4 & 190442  & +1.19 & 1620.51   & +1.88 & 0.007 \\
{5}          uncorrected        & 2/2 & {2.7}  & 4315 &       & 144435  &       & 3146.87   &       & 0.004 \\
                            {5} corrected         & 2/2 &                       & 5257 & +21.8 & 147948  & +2.43 & 3135.25   & -0.37 & 0.004 \\
{6}         uncorrected        & 2/2 & {4.5}  & 2433 &       & 253778  &       & 3996.95   &       & 0.003 \\
                           {6} corrected         & 2/2 &                       & 3075 & +26.4 & 259839  & +2.39 & 4010.57   & +0.34 & 0.003\\ \hline
\end{tabular}
\end{table*}

Table \ref{table:table7.3} presents the above metrics, together with the ratio of the aligned images by the total images, the average percentage of the pixels of the images containing caustics based on the predicted binary images, the number of the matched points (the number of the points of the sparse point cloud) and the respective percentage of the change achieved between the uncorrected and the corrected imagery, the number of the dense points and the respective percentage of change, the density D and the respective percentage of change and finally the roughness R.

By observing the metrics presented in Table \ref{table:table7.3}, a first important outcome is that when the corrected imagery is used, more images are aligned in the image alignment step. Moreover, it can be noticed that for all the performed tests, when the corrected imagery is used, more points are matched, delivering a denser sparse point cloud and a more robust 3D geometry of the scene. These first two outcomes were expected, and the background is already reported in Subsection \ref{subsection:section2.3.2}. The increase of the matched points, is proportional to the number of the images covered with caustics. However, there is not a strict relation between those two. This is not satisfied for test case 1 and test case 3, since not the same number of images are aligned. An increase in the number of the dense points is also observed. However, this is not of the same magnitude as that for the matched points. Again, here test case 1 and test case 3 are excluded for the same reasons as before. Coming now to the density of the point clouds and their roughness, no significant differences can be observed. 

Regarding the density, this was expected since the unaffected areas of the images are remaining the same, so there is no reason for generating more 3D points there. However, the fact that the roughness of the point clouds is remaining the same is of great importance; In the literature \cite{10} it is reported that most of the underwater image enhancement methods are increasing the roughness of the generated 3D point clouds. Nevertheless, this does not apply for the proposed methodology, highlighting the accuracy and the quality of the performed pixelwise image correction.

\section{Conclusion}
In this article, a novel method for correcting the radiometric effects of caustics on the shallow underwater imagery was presented. Additionally, R-CAUSTIC, the first real-world dataset on underwater rippling caustics was delivered and documented. 

The proposed method firstly relies on state-of-the-art deep learning tools which can accurately classify the pixels of the image as "non-caustics" or "caustics" and then exploits the 3D geometry of the scene in order to achieve a pixel-wise correction, by transferring color values between the overlapping images. The method depends on the good matches among the images, since an accurate fundamental matrix calculation is a prerequisite. However, if no good matches can be achieved, even when the detected masks are exploited in the key point detection step, images are not even appropriate for image-based 3D reconstruction, which this method intends to improve. 

Results suggest that the tested FCNs architectures achieve very high accuracy in this binary classification problem even from the first 15-20 epochs. Differences on the sensitivity of the models were also expected, however this highlighted the need to train a final model using all the available subsets. Pixel classification results over images captured for real world underwater photogrammetric applications in shallow waters suggested that the trained models can generalize over different types of seabed and caustics with high reliability. Caustics correction experiments performed illustrated the robustness and the reliability of the method over different types of seabed, different types of caustics and different anaglyph of the scene. The need of the color transferring step that is proposed was also highlighted.

Concerning the improvements on the 3D reconstruction of the scenes, the effectiveness of the proposed method was clearly obvious, since complete 3D point clouds were delivered, leaving no doubts about the achieved results. It was proven that when the corrected imagery is used for performing 3D reconstruction tasks, more images are aligned, and more points are matched. This delivers a more robust, complete and reliable 3D reconstruction. Moreover, it was also considered very important that the proposed method did not increase the roughness of the generated dense point clouds for all the testing cases. The proposed method will enable the users to capture less images and deliver more complete results, covering also larger areas. This will reduce the SfM-MVS processing time and the revisit of the underwater site for extra data will be avoided. 

By using the proposed method, a more chromatically consistent and realistic representation of the seabed is achieved. From the experiments performed, some remaining artifacts noticed on the images, especially in the areas having large radial distance, are attributed to the really intense chromatic aberration effect. However this was not to be dealt with within the context of this work. It was found that this effect is negatively affecting mainly the SfM-MVS process, compared with the skipping of the color transferring step. To overcome these issues, a channel-based correction of the refraction effect on the water-lens-air interface has to be performed. However, it can be easily avoided by using dome ports instead of flat ports and by avoiding fish-eye lenses and generally very small focal lengths. Although this approach was developed intending to correct the radiometrically affected areas on the underwater imagery, it can also be exploited to correct overwater, aerial and satellite imagery for specularities, shadows, and occlusions caused by illumination conditions, objects or even clouds.

\bibliographystyle{IEEEtran}
\bibliography{caustics}

\vfill

\end{document}